\documentclass[review]{elsarticle}

\usepackage{hyperref}
\usepackage{numcompress,amsmath,epsfig}
\usepackage{mathtools}
\usepackage{graphicx}
\graphicspath{{./fig/}}
\usepackage{subfigure}
\usepackage{booktabs}
\usepackage{multirow}
\usepackage{color}
\biboptions{sort&compress}
\journal{Journal of \LaTeX\ Templates}

\begin{document}

\begin{frontmatter}

\title{\textbf{TOV}: \textbf{T}he \textbf{O}riginal \textbf{V}ision Model for Optical Remote Sensing Image Understanding via Self-supervised Learning}

\author[address01]{Chao~Tao}
\author[address01]{Ji~Qi}
\author[address02]{Guo~Zhang}
\author[address03]{Qing~Zhu}
\author[address01]{Weipeng~Lu}
\author[address01]{Haifeng~Li\corref{mycorrespondingauthor}}
\cortext[mycorrespondingauthor]{lihaifeng@csu.edu.cn}
\address[address01]{School of Geosciences and Info-Physics, Central South University, Changsha, Hunan, PR China}
\address[address02]{State Key Laboratory of Information Engineering in Surveying, Mapping and Remote Sensing, Wuhan University, Wuhan, Hubei, PR China}
\address[address03]{Faculty of Geosciences and Environmental Engineering, Southwest Jiaotong University, Chengdu, Sichuan, PR China}


\begin{abstract}
Do we on the right way for remote sensing image understanding (RSIU) by training models via supervised data-dependent and task-dependent way, instead of human vision in a label-free and task-independent way? We argue that a more desirable RSIU model should be trained with intrinsic structure from data rather that extrinsic human labels to realize generalizability across a wide range of RSIU tasks. According to this hypothesis, we proposed \textbf{T}he \textbf{O}riginal \textbf{V}ision model (TOV) in remote sensing filed. Trained by massive unlabeled optical data along a human-like self-supervised learning (SSL) path that is from general knowledge to specialized knowledge, TOV model can be easily adapted to various RSIU tasks, including scene classification, object detection, and semantic segmentation, and outperforms dominant ImageNet supervised pretrained method as well as two recently proposed SSL pretrained methods on majority of 12 publicly available benchmarks. Moreover, we analyze the influences of two key factors on the performance of building TOV model for RSIU, including the influence of using different data sampling methods and the selection of learning paths during self-supervised optimization. We believe that a general model which is trained by a label-free and task-independent way may be the next paradigm for RSIU and hope the insights distilled from this study can help to foster the development of an original vision model for RSIU.
\end{abstract}


\end{frontmatter}


\section{Introduction}\label{sec:intro}

Human vision is nature capability with which one can easily perform remote sensing image understanding (RSIU) from coarse (scene) to fine (object) without any task-oriented learning. Modern RSIU has achieved remarkable progress based on machine vision model rather than human vision via teaching a machine to complete specific RSIU task, such as scene classification, object detection and semantic segmentation \cite{Cheng_Han_Lu_2017,Li_Dou_Tao_Wu_Chen_Peng_Deng_Zhao_2020,Shao_Yang_Zhou_2018,Xia_Bai_Ding_Zhu_Belongie_Luo_Datcu_Pelillo_Zhang_2018,luUnifiedDeepLearning2022}, through supervised training a task-specific model with human labeled task-specific data \cite{Zhou_Newsam_Li_Shao_2018,taoSpatialInformationInference2019,zhengHyNetHyperscaleObject2020}. Thus, compared what human vision do, a natural question is raised: Can this "teaching" method really solve the problem of RSIU?
In fact, this \"teaching\" way has many limitations in constructing visual models for RSIU:

1) Model training relies too much on large-scale, high-quality labeled data. When teaching machine to complete RSIU tasks, human label can be regarded as a kind of knowledge. Therefore, the more knowledge learned, the better the model performance \cite{Sun_Shrivastava_Singh_Gupta_2017,He_Girshick_Dollar_2019}. However, building a big remote sensing dataset is very challenging, as the accurate annotation of RSIs is tedious and require rich experience and geographic knowledge. Moreover, the annotating approach used for RSIU tasks is extremely task-dependent. For example, scene classification task requires image-level annotation while semantic segmentation task requires pixel-level annotation. Obviously, different labeling methods have different cost, which means a great amount of efforts need to be paid for constructing datasets in a task-dependent way.

2) More importantly, taking manual labels as supervised signals alone cannot learning a the desired vision model itself because the only function of manual labels, as extrinsic supervised signals, is to guide a model fitting to given training data. On the opposite, the intrinsic information hidden in massive remote sensing data should theoretically be much richer and more fundamental than the semantic information provided by human-labeled samples. Thus, human-labeled samples may be insufficient for annotating more complex scenes with multiple semantic meanings or with ambiguous semantic contents, which causes the problem of limited feature representation learning.

Unlike the machine vision that is "taught" by labeled data, human-like vision is achieved by holistic and joint models that can simultaneously solve real-world problems by unsupervised way \cite{Shao_Chen_Li_Wang_Yin_He_Teng_Sun_Gao_Liu_et}. The key reason is that human visual recognition system is not limited to a specific task or specific dataset, and human language based labels are not the prerequisite for constructing the human visual system. For example, a person who has never purposely learned any remote sensing knowledge can easily identify common objects such as farmland, vehicles and parking lots from optical remote sensing images. Inspired by this, we believe that training a general model in a label-free and task-independent way may be the next paradigm for RSIU that is closer to the human visual process.

Most recently, a new machine learning paradigm, self-supervised learning (SSL), has emerged in the field of natural language process (NLP) and computer vision (CV) \cite{Devlin_Chang_Lee_Toutanova_2019,Brown_Mann_Ryder_Subbiah_Kaplan_Dhariwal_Neelakantan_Shyam_Sastry_Askell_et,Jing_Tian_2021,Liu_Zhang_Hou_Mian_Wang_Zhang_Tang_2021}. Its main idea is to use human-designed task-agnostic self-supervised learning signals to generte pseudo-labels for massive unlabeled data, thereby replacing human label to guide the model learning. Since the model trained by self-supervised learning signals can be easily adapted to a wide range of downstream tasks, it can be considered as an general model. For example, original models like BERT \cite{Devlin_Chang_Lee_Toutanova_2019} and GPT-3 \cite{Brown_Mann_Ryder_Subbiah_Kaplan_Dhariwal_Neelakantan_Shyam_Sastry_Askell_et} have demonstrated significant effects on NLP. In the field of CV, researchers in Microsoft Research have built an fundamental or general vision model called Florence via unified Image-Text Contrastive Learning trained on Web-scale image-text data \cite{Yuan_Chen_Chen_Codella_Dai_Gao_Hu_Huang_Li_Li_et} and showed that the model trained from massive unlabeled data in a task-independent way can adapt to a wide range of downstream tasks such as classification, object detection, visual question answering, image caption, video retrieval, and action recognition.

In this paper, we forge the concept, The Original Vision (TOV) model, in the two dimensions of task modality and data granularity. For the task modality, TOV can be served as the start point to be modified to generalize across various RSIU tasks and RS modalities; For the data granularity, TOV can be served as the start point to pre-train a very big deep neural network with unlabeled massive scale RSIs which are extremely easy to obtain by Earth Observing System. Thus, As a new paradigm to overcome transfer catastrophe problem, TOV as a model which is trained on broad and massive data via self-supervision learning at very large can be adapted to wide diverse application tasks with very limited labeled samples. Pioneer works \cite{Tao_Qi_Lu_Wang_Li_2021,Li_Li_Zhang_Liu_Huang_Zhu_Tao_2022} have demonstrated surprising emergent capabilities on a wide range of downstream tasks and demonstrated the potential of SSL based machine learning paradigm on RSIU tasks. Seen these results, we argue that TOV models as a growing paradigm shift, where many applications will be directly derived from TOV models. However, there are still two key questions that remain unclear: 1) Are there general guidelines on creating benchmark datasets for TOV model learning and how to create them efficiently? 2) Under the SSL learning paradigm, how to train a high-performance TOV model from the perspective of model optimization?

This paper focuses on constructing TOV model from the perspective of task granularity, that is, training TOV using optical RSIs in a task-independent way but can generalize across various RSIU tasks. Then, for the first question, though SSL method can be used to learn TOV model without labels, we experimentally find that traditional grid sampling approach is not the best way on creating datasets for TOV model learning, since it may sample large amounts of semantically meaningless data. Besides, the classes in such dataset are severely imbalanced posing further challenges in SSL \cite{Li_Zhou_Xiong_Hoi_2020,Yang_Li_Huang_Liu_Hu_Peng_2021}. To solve this problem, we propose an automated data sampling and resampling mechanism guided by geographic data products like OSM and FROM-GLC10 to formulate a prototype for constructing massive, scalable and relatively balanced RSI dataset for training TOV model. By this way, we efficiently create two datasets collected from google earth, which contains 3 million class-imbalanced RSIs and 0.5 million relatively class-balanced RSIs, respectively. For the second question, though previous works have demonstrated the potential of using SSL for training TOV model in CV \cite{Goyal_Caron_Lefaudeux_Xu_Wang_Pai_Singh_Liptchinsky_Misra_Joulin_et}, there is a gap of model generalization compared to traditional supervised learning (e.g., ImageNet pretrained model) if it is applied in RSIU tasks. We argue that only using RSIs for TOV model learning may limit the discrimination and generalizability of the learned feature due to the characteristic of lower spatial-resolution and non-object centralization of RSI training samples. Thus, we propose a novel human-like SSL learning mechanism, which first learns general knowledge from web-scale natural images and then learns domain-relevant specialized knowledge from unlabeled RSIs. Experiments have shown that TOV model learned by the proposed method can adapt to various RSIU tasks (e.g., scene classification, object detection and semantic segmentation) and achieves state-of-the-art results in majority of 12 publicly available RSIU benchmarks. The contributions of this paper are twofold:

1) We firstly define the TOV model for RSIU and analyze the influences of using different data sampling methods and the selection of learning paths during self-supervised optimization.

2) We release the benchmark dataset for training the TOV model as well as the pretrained model\footnote{\href{https://github.com/GeoX-Lab/G-RSIM/tree/main/TOV\_v1}{https://github.com/GeoX-Lab/G-RSIM/tree/main/TOV\_v1}}, which can help to foster the development of TOV model in the remote sensing community.

3) We propose a novel human-like SSL learning mechanism inspired by the followed insights: only using RSIs for TOV model learning may limit the discrimination and generalizability of the learned feature due to the characteristic of lower spatial resolution and non-object centralization of RSI training samples.

4) Experiments have shown that the TOV model learned by the proposed method can adapt to various RSIU tasks (e.g., scene classification, object detection, and semantic segmentation) and achieves state-of-the-art results in the majority of 12 publicly available RSIU benchmarks. To our best knowledge, this is the first work to successfully train a general model, called TOV, by million-level label-free samples and achieve state-of-the-art results for various RSIU tasks simultaneously. 

The remainder of this paper is organized as follows: Section \ref{sec:rela} gives the definition of TOV model for RSIU. The details of constructing TOV model are presented in Section \ref{sec:method}. The generalization performance of TOV model on various RSIU tasks is evaluated in Section \ref{sec:exp}. Finally, discussions and conclusions are presented in Section \ref{sec:dis} and \ref{sec:conclusion}, respectively.

\begin{figure*}[!t]
  \centering
  \includegraphics*[width=0.85\linewidth]{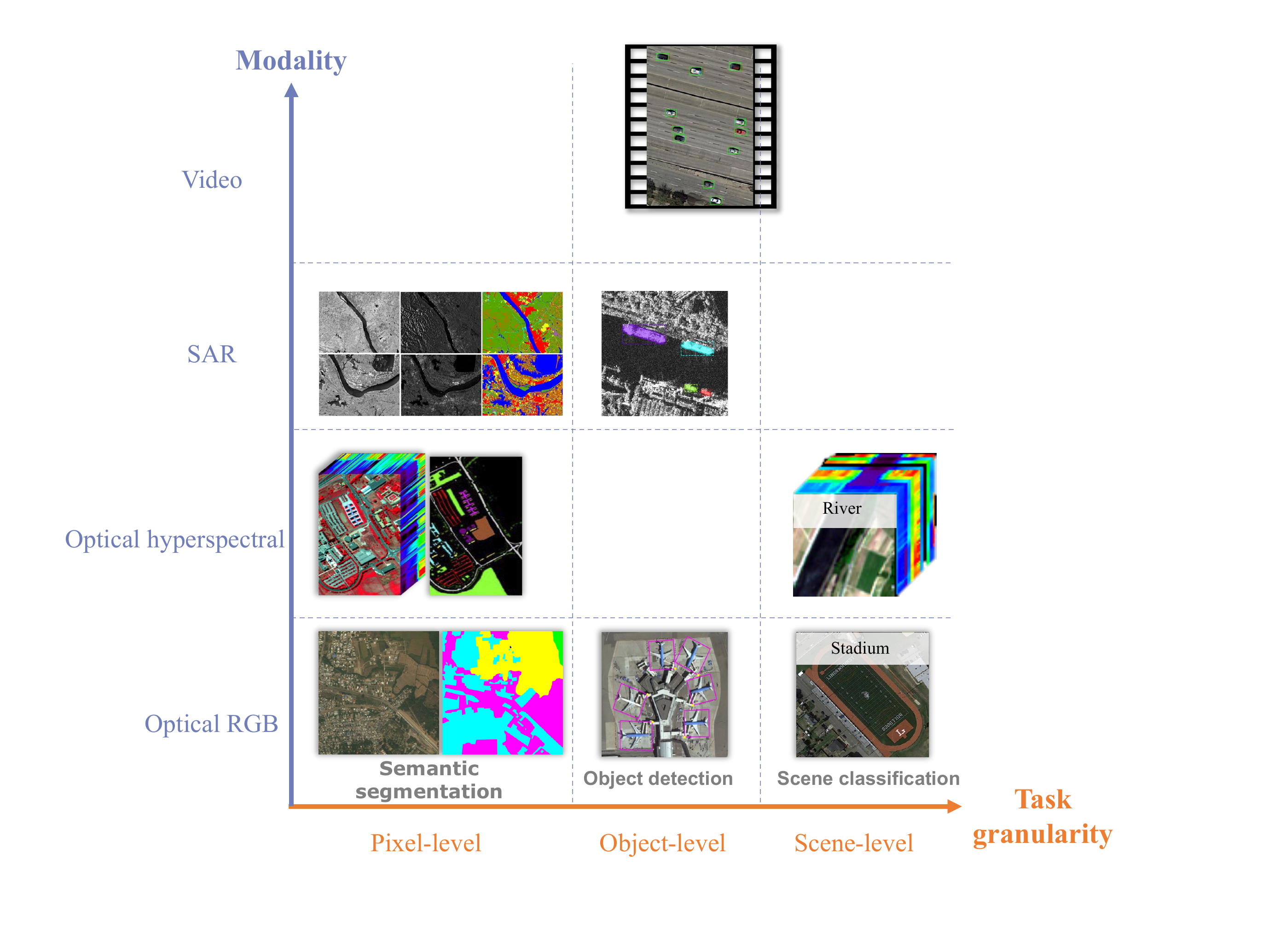}
  \caption{The problem space of RSIU containing task granularity and data modality dimensions.}
  \label{fig:TOV_definition}
\end{figure*}  

\section{The definition of TOV model for RSIU}\label{sec:rela}
Although some original models, like GPT-3 \cite{Brown_Mann_Ryder_Subbiah_Kaplan_Dhariwal_Neelakantan_Shyam_Sastry_Askell_et} and Florence \cite{Yuan_Chen_Chen_Codella_Dai_Gao_Hu_Huang_Li_Li_et}, have demonstrated good performance in NLP and CV, the concept of TOV model is new in the field of remote sensing. Thus, in this section, we firstly define TOV model for RSIU from the two dimensions of task granularity and data modality.

From the perspective of task granularity, RSIU tasks can be divided into three levels: scene-level, object-level and pixel-level corresponding to understand RSIs from coarse to fine. From the perspective of data modalities, RSIs can be divided into four modal data types, including optical (multispectral, hyperspectral), satellite video, and SAR, according to the imaging mechanism of remote sensing data and the difference in spectral sensing range. Thus, as shown in Figure \ref{fig:TOV_definition}, RSIU tasks can be mapped to a problem space containing two dimensions of task granularity and data modality, and TOV model should be a general vision solution for all tasks in the above problem space. With this motivation, we define TOV model for RSIU to be a pre-trained model trained in a task-independent and modality-independent way but can easily adapt to (e.g., fine-tuned) a wide range of RSIU tasks and data modalities.

Considering the complexity of the problem, this paper focuses on constructing TOV model from the perspective of task granularity.

\section{Methodology}\label{sec:method}
\subsection{The overall framework for constructing TOV model for RSIU}\label{itm:overall}
\begin{figure*}[!t]
  \centering
  \includegraphics*[width=\linewidth]{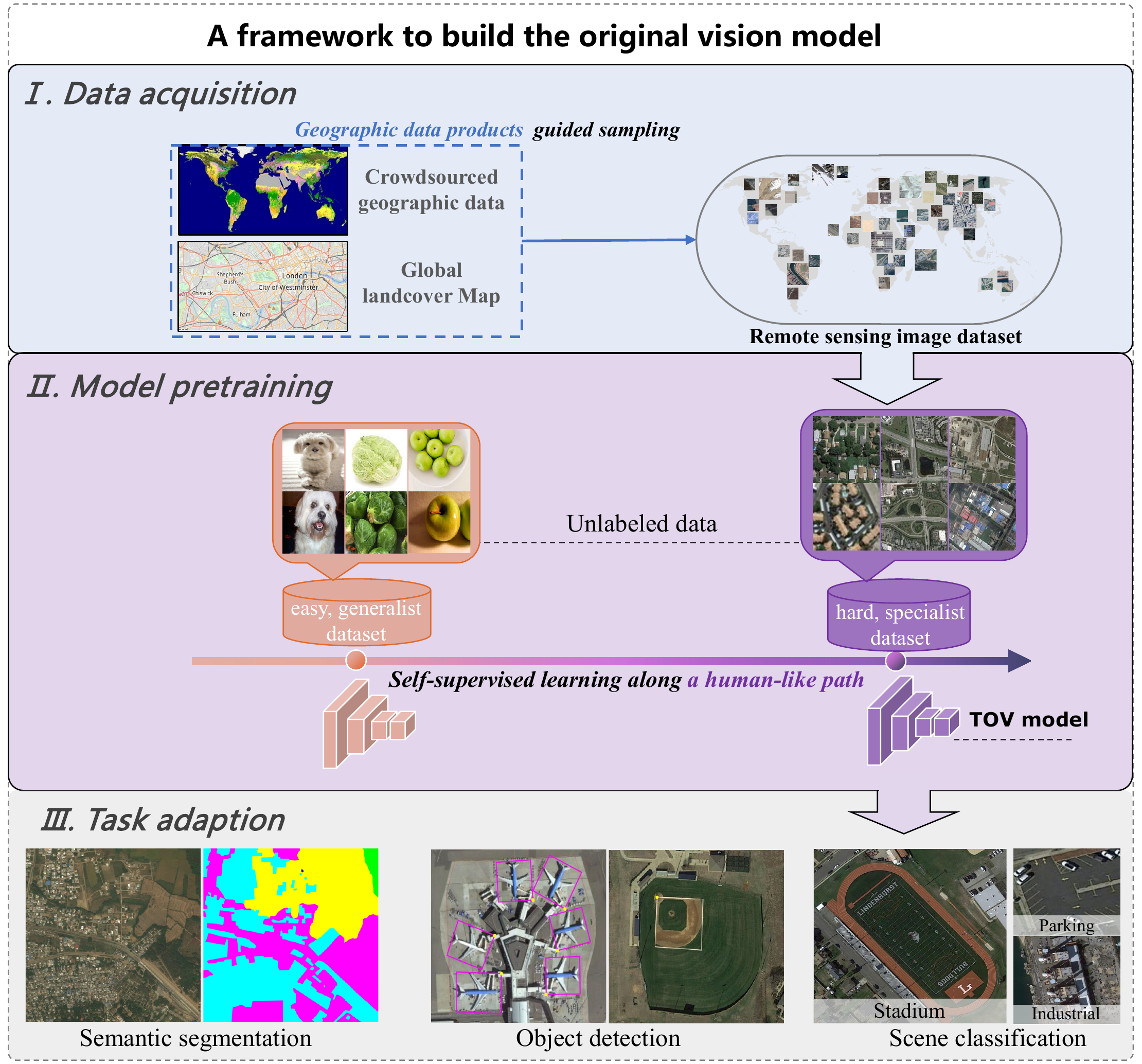}
  \caption{Overview of building TOV model for RSIU.}
  \label{fig:TOV_framework}
\end{figure*}

Constructing TOV model for RSIU includes three stages: data acquisition, model pretraining and task adaptation, as shown in Figure \ref{fig:TOV_framework}.

\textbf{Data acquisition.} Rich and high-quality data is important for training TOV model. Although this model can be trained in a label-free SSL way, it can hardly learn valuable remote sensing visual knowledge from an unlabeled dataset that contains large amounts of semantically meaningless content or has severely class-imbalanced distribution. To address this problem, we propose an automatic RSIs sampling and resampling method (Sec. \ref{subsec:data_acquisition}) guided by public geographic data products, which can automatically collect relative class-balanced RSI samples with rich semantic content over global scale at low cost.

\textbf{Model pretraining.} Visual representation that generalizes well to different RSIU tasks should be both discriminative enough to train a strong classifier and invariance to significantly varying imaging conditions \cite{Qi_Luo_2020}. To achieve this, we propose a novel human-like SSL learning mechanism under the contrastive learning framework \cite{Chen_Kornblith_Norouzi_Hinton_2020} for training TOV, which first learns general knowledge from web-scale natural image dataset and then learns domain-relevant specialized knowledge from the constructed RSI dataset (Sec. \ref{subsec:ssl_mechanism}).

\textbf{Task adaption.} TOV is expected to be adaptable to various RSIU tasks. Therefore, the learned representations are first stored as parameters in TOV model. Then, we adapt the learned general representation from TOV model to various RSIU tasks by adding task-specific adapters following the backbone of TOV model. Specifically, a simple fully-connected layer is used as an adapter for scene-level understanding tasks; region proposal network and ROIHead in Faster RCNN (i.e., a popular two-stage object detection model) \cite{Ren_He_Girshick_Sun_2017} are jointly used as adapters for object-level understanding tasks; the common decoder in Fully Convolutional Networks (FCN) \cite{longFullyConvolutionalNetworks2015} is used as an adapter for pixel-level understanding tasks. In this way, TOV can effectively adapt to RSIU tasks with few labeled data.

\begin{figure*}[!t]
  \centering
  \includegraphics*[width=0.98\linewidth]{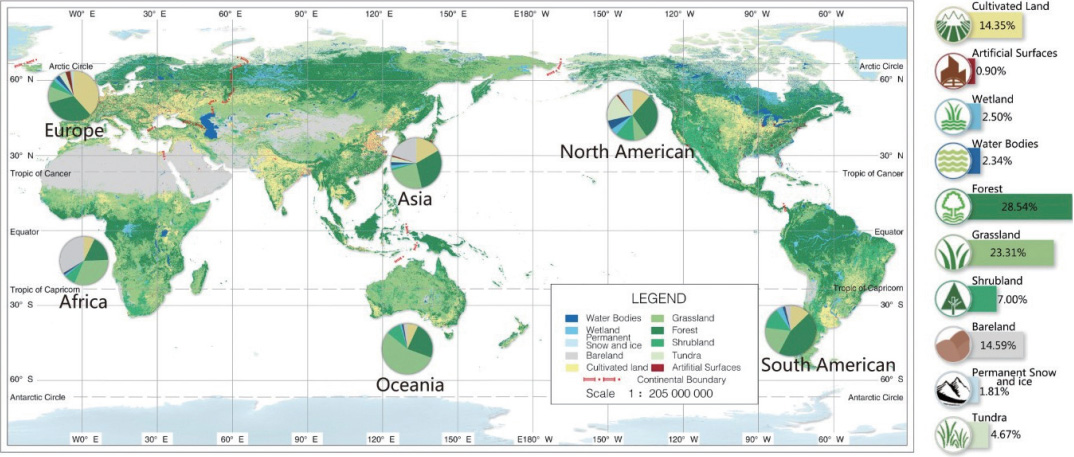}
  \caption{Statistical results of global land cover geographic elements based on GlobeLand30 2010. Copyright by National Geomatics Center of China.}
  \label{fig:glc_static}
\end{figure*}  

\subsection{Data acquisition for TOV model learning}\label{subsec:data_acquisition}
Compared with supervised learning methods, SSL methods can learn visual knowledge from unlabeled dataset, which provides a way to build TOV model for RSIU at low cost. But whether models can learn valuable visual knowledge depends on the semantic content richness and class balance of the samples in the unlabeled dataset \cite{Cole_Yang_Wilber_Mac,Tao_Qi_Lu_Wang_Li_2021}. Therefore, the unlabeled RSI dataset used for TOV model learning should have the above two key properties.

\begin{figure*}[!t]
  \centering
  \includegraphics*[width=0.98\linewidth]{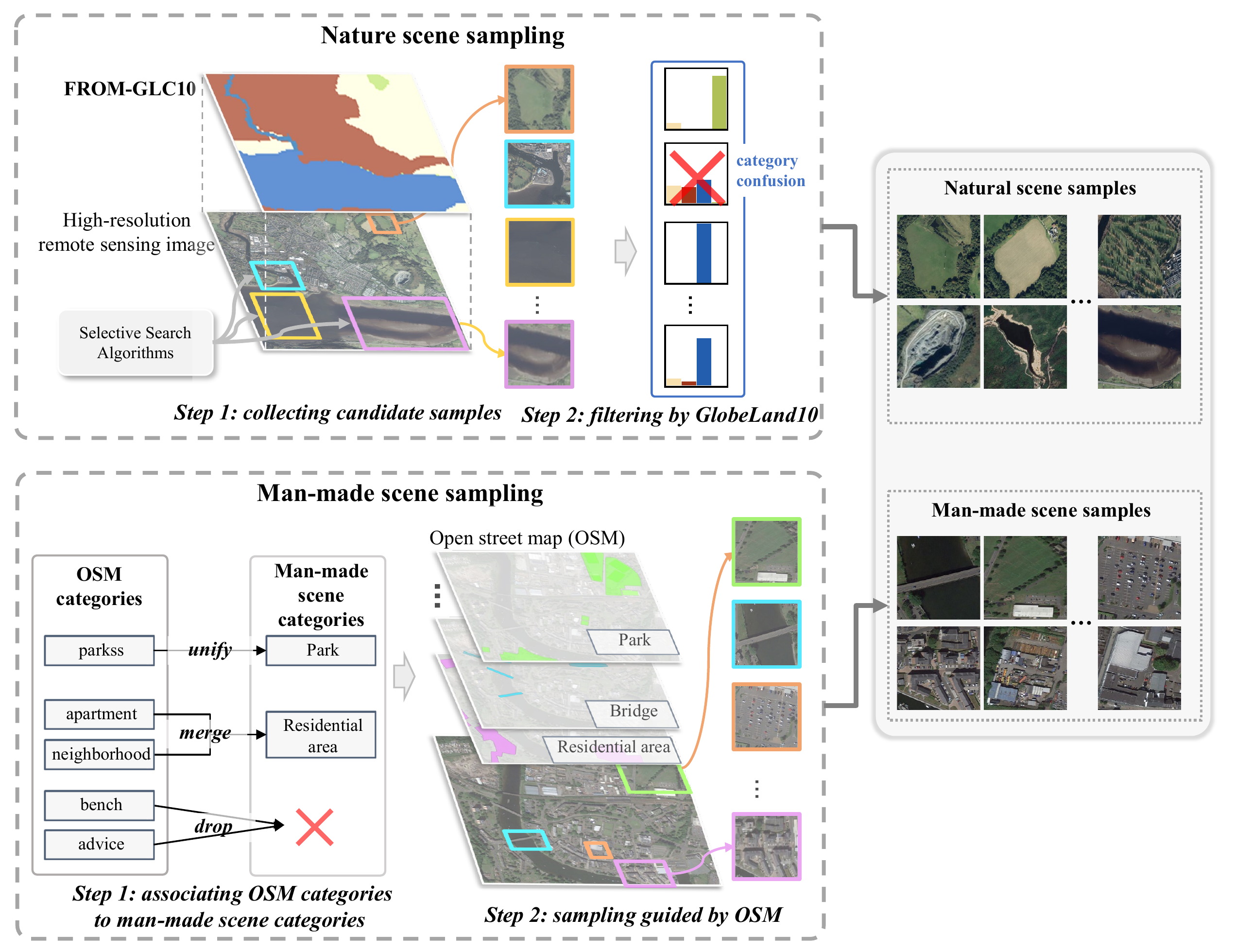}
  \caption{Flowchart of automatic sampling methods for natural and man-made scenes guided by geographic data product.}
  \label{fig:data_sampling}
\end{figure*}  

Without too much human involvement, constructing an RSI dataset with these two attributes is a challenge task. Though the traditional grid sampling approach is simple and straightforward, it may sample large amounts of semantically meaningless data. Besides, the classes in such dataset are usually severely imbalanced. The main reasons are: First, geographic elements are naturally severely unbalanced in quantity at global scale, as shown in Figure \ref{fig:glc_static}\footnote{https://www.webmap.cn/commres.do?method=globeDetails\&type=GeographicalStatistics} \cite{Jun_Ban_Li_2014}. Second, the geographic elements shown in RSIs usually cannot be completely divided by the uniform grid. As a result, many samples contain mixed geographic elements making it even more impossible to control the class balance of the dataset. To solve these problems, we propose an automatic RSIs sampling and resampling method guided by the geographic data product to efficiently build a large-scale RSI sample dataset.

\begin{table}[!t]
  \centering
  \caption{Natural scene categories.}
  \resizebox{\textwidth}{!}{
    \begin{tabular}{cl}
      \toprule
      \textbf{Name} & \textbf{Description} \\
      \midrule
      Forest & A large area of land that is thickly covered with trees \\
      Grassland & A large area of open land covered with wild grass. \\
      Shrubland & Land on which shrubs are the dominant vegetation. \\
      Cropland & Land that is suited to or used for crops. \\
      Wetland & Land covered by swamps or marshes. \\
      Water & An area of water, especially a lake, river, sea or ocean. \\
      Tundra & Land where no trees grow and the soil below the surface is always frozen. \\
      Bareland & Land covered with no vegetation or buildings. \\
      Snow/Ice & Land covered by snow or ice. \\
      \bottomrule
      \end{tabular}
  } \label{tab:categories_of_GLC}
\end{table}  

Geographical elements can be divided into natural scene elements (e.g., forests, meadows, water bodies) and man-made scene elements (e.g., residential areas, industrial areas, schools, parking lots). These two types of geographic elements differ significant in spatial distribution, temporal distribution, and the corresponding geographic data product forms (e.g., global land cover map and crowdsourced geographic data). Thus, we firstly design two sampling methods to respectively collect samples of these two categories (as shown in Figure \ref{fig:data_sampling}). Then, the resulting dataset is resampled using the noise label provided by the geographic data product to obtain a relatively class-balanced dataset.

\subsubsection{Data sampling for natural scene elements}\label{subsubsec:sampling_natural}

Considering the characteristics of low scale variance and slow temporal change of natural scene elements in RSIs, we use the global land cover mapping product with the spatial resolution of 10 m (FROM-GLC10 \cite{Gong_Liu_Zhang_Li_Wang_Huang_Clinton_Ji_Li_Bai_et}) to guide the sampling of natural geographical elements shown in Table \ref{tab:categories_of_GLC}. The sampling process consists of the following two steps:

Step 1: Automatic collection of candidate samples. To avoid the redundant and invalid sampling existing in traditional grid sampling methods, in this step, we utilize the region proposal mechanism that is commonly used in object detection, where the classic selective search algorithm \cite{uijlingsSelectiveSearchObject2013} is used to select candidate regions from the input image for collecting candidate samples. Specifically, each image $I_i$ in the dataset $H=\{I_{1}, I_{2} \ldots I_{N_{I m g}}\}$ is firstly over-segmented into super-pixels by a graph-based image segmentation method \cite{Felzenszwalb_Huttenlocher_2004}. Then, adjacent small super-pixels are continuously grouped based on the similarity metric established by color, texture, and shape features. In this way, image $I_i$ can be segmented into $n_i$ segments. Finally, $N_1=\sum^{N_{Img}}_{i=1}{n_i}$ candidate samples are collected by taking the minimum bounding rectangle of each segment.

Step 2: Sample selection guided by FROM-GLC10. To facilitate the subsequent resampling process (see Sec. \ref{subsubsec:resampling}), we prefer each sample containing a single type of natural scene elements to avoid category confusion. Therefore, we establish the following metric to evaluate the category homogeneity of each candidate sample:
\begin{equation}
\label{equ:equ_score}
S_{i}=\sum_{c=1}^{C_{nautre}} p_{i, c} \log (p_{i, c})
\end{equation}
where larger $S_i$ means that the proportion of a type of natural scene elements contained in this sample is much higher than other natural scene elements. $p_{i,c}$ denotes the percentage of pixels of class $c$ in the $i$-th candidate sample. And $C_{nautre}$ represents the number of categories as shown in Table \ref{tab:categories_of_GLC}. We score the candidate samples by Eq. (\ref{equ:equ_score}), and drop those with scores less than $T$. The value of $T$ is empirically set as 0.2 to trade off the category richness and homogeneity of the obtained samples.

Through above steps, we obtain a natural scene dataset $S_{nautre}$ with $N_{nautre}$ samples.

\begin{table}[!t]
  \centering
  \caption{The rule to associate OSM categories to man-made scene categories.}
  \resizebox{\textwidth}{!}{
    \begin{tabular}{cc}
      \toprule
      \textbf{Scene categories} & \textbf{OSM categories} \\
      \midrule
      Airport & aerodrome, airfield, apron, security, aerohangar, waiting aera, terminal, hangar, … \\
      Parking & parking, disused parking, parking space, car park, parking, car pooling, … \\
      Commercial area & shopping center, retail, marketplace, wholesale, commercial, shopping mall, … \\
      Residential area & residential, apartment, terrace, terrace house, townhouse, neighborhood, … \\
      School & university, college, school, secondary school, education, … \\
        … & … \\
      Sports center & volleyball, sport, playground, netball, court, playing field, recreation ground, … \\
      \bottomrule
    \end{tabular}
  } \label{tab:categories_of_OSM}
\end{table}  

\subsubsection{The automatic sampling method for man-made scene elements}\label{subsubsec:sampling_man_made}
Considering the high scale variance and rapid temporal change of man-made scene elements in RSIs, we adopt frequently updated and fine-grained Open Street Map (OSM)\footnote{https://www.openhistoricalmap.org} to guide the sampling of man-made geographical elements. The sampling process has the following two steps:

Step 1: Associating OSM categories to man-made scene categories. OSM contains many redundant categories and invalid categories. Redundant categories have similar semantic contents (e.g., "college" and "university"), which prevents obtaining a balanced category distribution during resampling (see Sec. \ref{subsubsec:resampling}). Invalid categories (e.g., "phone", "advice") may be unrelated to geographical element, which may prevent providing proper guidance for collecting desired samples. To address these issues, we firstly construct a man-made scene classification system with $C_{man-made}$ categories \cite{Cheng_Han_Lu_2017,Christie_Fendley_Wilson_Mukherjee_2018,Li_Dou_Tao_Wu_Chen_Peng_Deng_Zhao_2020,Long_Xia_Li_Yang_Yang_Zhu_Zhang_Li_2020,Xia_Hu_Hu_Shi_Bai_Zhong_Zhang_Lu_2017}. Then, we define a mapping rule shown in Table \ref{tab:categories_of_OSM} to associate OSM categories to man-made scene categories.

Step 2: Automatic sampling guided by OSM. Considering that the man-made scene elements may change rapidly over time, we firstly retrieve OSM data $M_i$ that matches the input image $I_i$ in space and time as much as possible for sampling. Then, we parse $n_i$ man-made scene elements from $I_i$ based on the crowd-sourced annotations in $M_i$ and the association rules defined in step 1. Afterwards, we determine the sampling region of each element in $I_i$ by geographic coordinate transformation, and finally collect $n_i$ samples by taking the minimum bounding rectangle of each region.

Through above step, we obtain a man-made scene dataset $S_{man-made}$ with the size of $N_{man-made}=\sum^{N_{Img}}_{i=1}{n_i}$.

\subsubsection{Class-balanced oriented resampling strategy}\label{subsubsec:resampling}
Many recent studies have investigated SSL in the context of class-imbalanced, and consistently observed the undesired performance of existing SSL algorithm \cite{Wei_Sohn_Mellina_Yuille_Yang_2021}. Thus, constructing a class-balanced dataset is very important for promoting the performance of TOV model for RSIU even it is trained with label-free dataset.

After the above two sampling methods, we construct an RSI dataset with \linebreak 3 million samples of 31 categories, and name it as TOV-RS-imbalanced. However, this dataset is not class-balanced since the distribution of geographical elements is naturally uneven in space. To alleviate this problem, we develop a class-balanced oriented resampling strategy, which contains the following two steps:

1) Finding the class with the least sample data from the natural scene feature dataset $S_{nature}$, and let $n_k$ represent the number of samples in this class. Then, we randomly choose $n_k$ samples from each class in $S_{nature}$, and construct a relatively class-balanced subset $S_{nature}^{\prime}$ containing $n_k\times C_{nature}$ samples.

\begin{figure*}[!t]
  \centering
  \includegraphics*[width=0.98\linewidth]{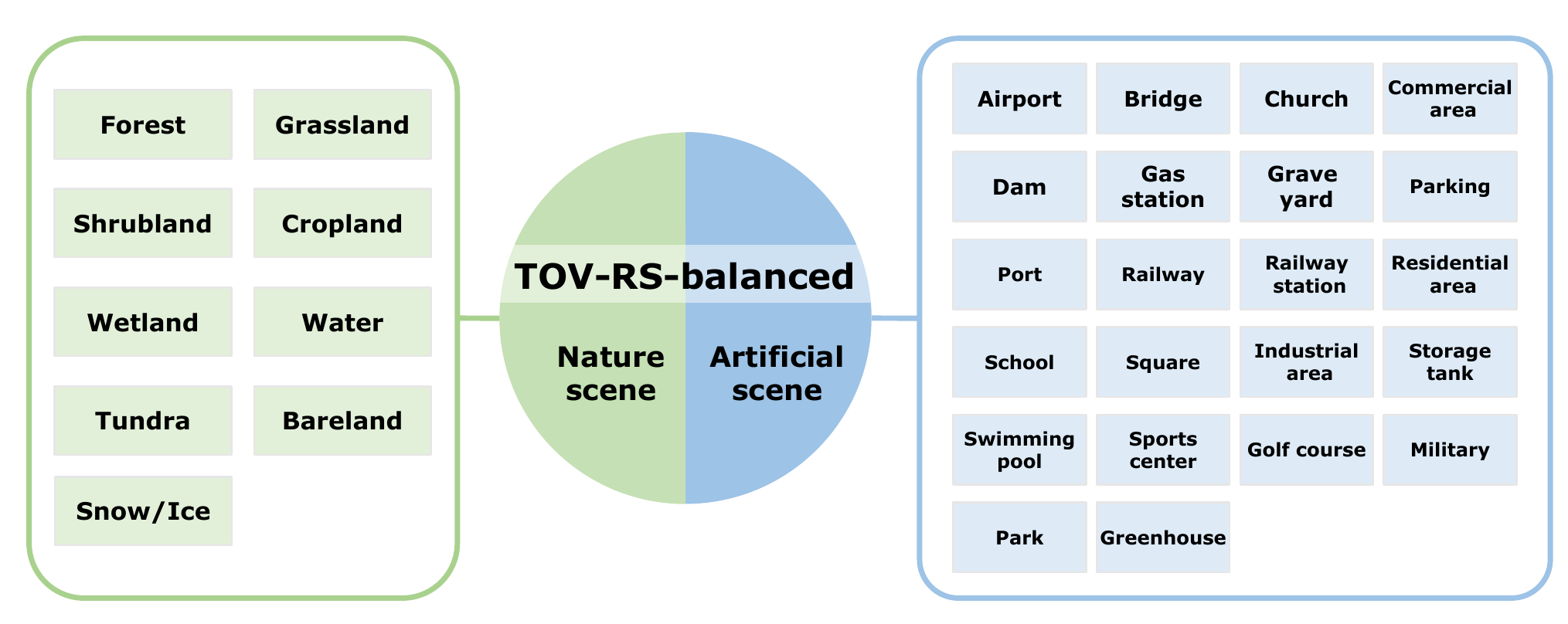}
  \caption{Semantic types of samples in the constructed TOV-RS-balanced dataset.}
  \label{fig:tov_rs}
\end{figure*}

2) Similarly, a relatively class-balanced subset $S_{man-made}^{\prime}$, containing $n_{k}^{\prime} \times C_{man-made}$ man-made scene samples ($n_{k}^{\prime}=\frac{n_{k} \times C_{nautre}}{C_{man-made}}$), is obtained from \linebreak $S_{man-made}$. Finally, combing $S_{nature}^{\prime}$ and $S_{man-made}^{\prime}$, we obtain a relatively class-balanced dataset, TOV-RS-balanced, with 0.5 million samples of 31 categories. Details about the dataset TOV-RS-balanced is shown in Figure \ref{fig:tov_rs}.

To sum up, the proposed automated data sampling and resampling method can construct dataset with three notable characteristics, which contributes to training TOV model for RSIU:

1) Diversity: the samples in TOV-RS-balanced are collected in the categories from nature geographic elements and man-made geographic elements, with the spatial resolutions between 1-20 m, regions covering more than 100 countries and time phases from 2019-2021, so they are diverse in category, spatial resolution, illumination, and background.

2) Relatively class-balanced: TOV-RS-balanced can maintain the balance of samples from natural scenes and man-made scenes, because we use two different methods to sample such two kinds of scenes. But it can only achieve relatively class balance in sub-categories due to the noisy annotations in geographic data products.

3) Scalable: TOV-RS-balanced can expand in category, quantity of objects, and diversity since the proposed RSIs sampling and resampling is automated. Moreover, its data source can also extend to other data modalities of RSI.

\subsection{Training TOV model for RSIU based on a human-like SSL mechanism}\label{subsec:ssl_mechanism}
Recent studies have demonstrated that TOV model trained by contrastive self-supervised learning with mass unlabeled nature images has impressive generalizability, which perform comparably well or even better than supervised learning methods across various computer vision tasks \cite{Yuan_Chen_Chen_Codella_Dai_Gao_Hu_Huang_Li_Li_et,Goyal_Caron_Lefaudeux_Xu_Wang_Pai_Singh_Liptchinsky_Misra_Joulin_et,Shao_Chen_Li_Wang_Yin_He_Teng_Sun_Gao_Liu_et}.
However, we experimentally find that directly using this pipeline to train TOV model for RSIU cannot obtain desired results. The main reasons could be: 1) Compared with natural images, RSIs have lower spatial resolution and blurrier texture details, from which learning low-level visual knowledge (e.g., texture features and edge features) is difficult. 2) Most RSI samples contain a complex scene consisting of multiple ground objects, which makes self-supervised optimization more difficult.

Human usually learn knowledge along a path from easy to difficult and from general to specialized. We refer such path as the human-like learning path. For example, a person who has never purposely learned any remote sensing knowledge can identify common geographic elements in the real world like buildings, roads and vehicles, but he may need long-term remote sensing knowledge learning to identify some complex geographic elements such as tundra, coniferous forest, and broadleaf forest. These phenomena inspire us that: 1) Common knowledge exists between RSIs and natural images, and the general low-level visual representations that are difficult to learn from RSIs can be learned from natural images; 2) Since the knowledge for RSIU is more complex and specialized than that of natural image understanding, it is better to build TOV model for RSIU along an easy to difficult and general to specialized learning path, similar as what human do. With this motivation, we propose a human-like SSL mechanism to learn TOV model for RSIU. In the following, we first briefly introduce the contrastive SSL framework, and then detail the proposed human-like SSL learning path.

\subsubsection{Self-supervised contrastive learning}\label{subsubsec:sscl}
Contrastive learning is a typical SSL method, and it outperforms other unsupervised learning methods in learning general feature representation \cite{Chen_Kornblith_Norouzi_Hinton_2020,He_Fan_Wu_Xie_Girshick_2020,Tian_Krishnan_Isola_2019}. Contrastive learning methods bring different augmented views (positive sample pairs) of the same image closer and separate views (negative sample pairs) of different images, to learn both invariant and distinguishable visual representation. Specifically, it consists of the following two steps:

Step 1: Given a training set $X=\{\boldsymbol{x}_{1}, \boldsymbol{x}_{2}, \ldots, \boldsymbol{x}_{n}\}$ containing $n$ unlabeled samples, each sample ${\boldsymbol{x}}_i$ is augmented by $T(\cdot)$ to create two views as a pair of positive samples (${\boldsymbol{x}}^1_i\boldsymbol{,\ }{\boldsymbol{x}}^2_i$). In contrast, any two augmented views of different samples are treated as a negative sample pair $({\boldsymbol{x}}^1_i\boldsymbol{,\ }{\boldsymbol{x}}^2_j$). Here, $T(\cdot)$ is a stochastic set of augmentations including random crop, random flip, color distortion and Gaussian blur.

Step 2: Training a model to distinguish the positive and negative samples by embedding them to a proper feature space $f(\cdot)$ using the loss function defined as Eq. \eqref{equ:equ_sscl}.
\begin{equation}\label{equ:equ_sscl}
\footnotesize
\mathcal{L}=-\mathrm{E}_{X}\left[\log \frac{\exp \left(g\left(f\left(\boldsymbol{x}_{i}^{1}\right)^{T} f\left(\boldsymbol{x}_{i}^{2}\right) / \tau\right)\right)}{\exp \left(g\left(f\left(\boldsymbol{x}_{i}^{1}\right)^{T} f\left(\boldsymbol{x}_{i}^{2}\right) / \tau\right)\right)+\sum_{j=1}^{n-1} \exp \left(g\left(f\left(\boldsymbol{x}_{i}^{1}\right)^{T} f\left(\boldsymbol{x}_{j}^{2}\right) / \tau\right)\right)}\right]
\end{equation}
where $g(\cdot )$ is a multilayer projection head with 2 fully connected layers, which is widely used in SSL to compress the extracted features for contrasting. $\tau = 0.5$ is a temperature scalar. By minimizing Eq. \eqref{equ:equ_sscl}, positive samples are pulled closer while negative samples are pushed apart in the learned feature space, which enhances the invariance and distinguishability of the learned visual representation.

\subsubsection{Training TOV model for RSIU along a human-like SSL mechnism}\label{subsubsec:human_like_learning_path}
The proposed human-like SSL mechanism consists of the following two stages:

Stage 1: Learning general visual knowledge from natural image dataset. Since natural images have higher resolution and richer texture details than RSIs, we first perform self-supervised contrastive learning by using large-scale unlabeled natural images for learning general low-level visual features such as textures and edges. Specifically, we first construct a web-crawled natural image dataset containing 1 million samples. Then we train TOV model using the loss function defined by Eq. \eqref{equ:equ_sscl} in an SSL way. The learned model is denoted as $f_{1}(X; \boldsymbol{\rm W}_{B})$, where the learned general low-level visual knowledge is stored in weights ${\boldsymbol{\rm W}}_{B}$.

Stage 2: Learning specialized visual knowledge from the constructed RSI dataset. A common idea to achieve this is to initialize the model using the parameters learned in stage 1 and perform a secondary learning by using the RSI dataset. However, this approach may surfer the problem of catastrophic forgetting \cite{Goodfellow_Mirza_Xiao_Courville_Bengio_2015}, which occurs specifically when the network is trained sequentially on different datasets or tasks. To better connect these two learning stages, we designed a simple memory retention strategy, which fix the weights of the shallow and middle layers of the network learned in stage 1 to keep the memory of low-level visual representations and then use RSIs dataset to optimize weights of other layers of the network:
\begin{equation}
\min _{\{w \mid w \notin \boldsymbol{\rm W}_{b} \& w \in \boldsymbol{\rm W}_{b}^{\prime}\}} \mathcal{L}
\end{equation}
where $\mathcal{L}$ is the self-supervised contrastive learning loss defined in Eq. \eqref{equ:equ_sscl}. $\boldsymbol{\rm W}_b$ represents the weights fixed in stage 2, $\boldsymbol{\rm W}^{\mathrm{'}}_b$ represents the weights needed to be optimized and ${\boldsymbol{\rm W}}_b+{\boldsymbol{\rm W}}^{\mathrm{'}}_b={\boldsymbol{\rm W}}_B.$ The selection of ${\boldsymbol{\rm W}}_b$ and ${\boldsymbol{\rm W}}^{\mathrm{'}}_b$ is based on the studies about the properties of the learned visual representation of each layer in deep convolutional neural network \cite{Qin_Yu_Liu_Chen,Yosinski_Clune_Bengio_Lipson_2014}, that is, the layers close to the input layer tends to learn the general low-level feature representation, while the layers distant from input layer tends to learn specific high-level feature representation.

\section{Experiments}\label{sec:exp}

\subsection{Datasets for training TOV model}\label{subsec:pretraining_dataset}
To train TOV model along a learning path from general knowledge to specialized knowledge, we construct two datasets TOV-NI and TOV-RS respectively.

TOV-NI is a web-crawled natural image dataset of 1 million samples. It is used for learning general knowledge in TOV model. To cover a set of visual concept as broad as possible, we automatically search images from the internet using 10,000 text queries from Wordnet \cite{Fellbaum_Miller_1998}. Then, we choose at most 100 images per query to approximately keep the class balance of the resulting dataset.

TOV-RS is an RSI dataset constructed by the proposed data sampling method, which is used for learning specialized knowledge in TOV model. It has two versions. One is TOV-RS-imbalanced containing 3 million class-imbalanced samples and the other is TOV-RS-balanced containing 0.5 million class-balanced samples.

\subsection{Comparsion experiments and baseline}
In the experiments, we evaluate the performance and generalization capabilities of TOV model by three kinds of downstream RSIU tasks on 12 publicly available benchmarks, including scene classification, object detection and semantic segmentation. We also compared the proposed method with two commonly used model initialization methods and two recently proposed SSL methods, including:

1) Random initialization. Directly train a Resnet-50 model with random initialization parameters for each downstream RSIU task.

2) ImageNet pretraining. Directly train a Resnet-50 model initialized by the parameters pretrained on ImageNet \cite{Russakovsky_Deng_Su_Krause_Satheesh_Ma_Huang_Karpathy_Khosla_Bernstein_et} for each downstream RSIU task.

3) SimCLR \cite{Chen_Kornblith_Norouzi_Hinton_2020}. A self-supervised pretraining method used to train TOV model. In SimCLR, the transformed instances of the same sample are regarded as positive instances and that of different samples in a training batch are regarded as negative instances. The model learns visual knowledge by enhancing the similarity of positive instances and difference of negative instances.

4) MoCov2 \cite{He_Fan_Wu_Xie_Girshick_2020, Chen_Fan_Girshick_He_2020}. MoCov2 is also a self-supervised pretraining method used to train TOV model. Similar as SimCLR, it is also based on contrastive learning, but it focuses on obtaining negative instances with the size far beyond the batch size to learn more discriminative visual knowledge. Therefore, a dynamic queue is proposed to save the features of negative samples, and a momentum update encoder is proposed to avoid the consistency problem of the representations of negative samples from the rapid change of the encoder.

\subsection{Implementation and training detail}
In the experiments, we use Resnet-50 as the backbone to train TOV model for both the proposed method and two comparison SSL methods. The proposed method is trained using TOV-NI dataset and TOV-RS dataset continually along a human-like learning path while two compared SSL methods are trained only using the TOV-RS dataset. For all the three methods, we use the Adam optimizer with a batch size of 1024. The learning rate was initially set as 0.75 and was reduced in a cosine manner within 800 epochs. All experiments were implemented in PyTorch environment under the CentOS 7.5 platform with 8 NVIDIA Tesla A100 (memory 32 GB).

\subsection{Results and analysis}\label{subsec:results}

\subsubsection{Scene classification}\label{subsubsec:secene_classification}
\begin{table}[!t]
\normalsize
  \centering
  \caption{Datasets used for scene classification experiments.}
  \resizebox{\textwidth}{!}{
    \begin{tabular}{ccccccccc}
    \toprule
    \multirow{2}[4]{*}{Dataset} & \multicolumn{5}{c}{High-resolution RSIs datasets} &  & \multicolumn{2}{c}{Multi-spectral RSIs datasets} \\
\cline{2-6} \cline{8-9} \multicolumn{1}{c}{} & AID   & NR    & RSD46 & PatternNet & \multicolumn{1}{c}{UC Merced} & & EuroSAT & NaSC-TG2 \\
    \midrule
    Number of categories & 30 & 45 & 46 & 38 & 21 & & 10 & 10 \\
    Number of samples & 10,000 & 31,500 & 117,000 & 30,400 & 2,100 & & 27,000 & 20,000 \\
    Spatial resolution (m) & 0.5$\sim$8 & 0.2$\sim$30 & 0.5$\sim$2 & 0.062$\sim$4.693 & 0.3 & & 10$\sim$60 & 100 \\
    Image sizes & 600$\times$600 & 256$\times$256 & 256$\times$256 & 256$\times$256 & 256$\times$256 & & up to 64$\times$64 & 128$\times$128 \\
    \bottomrule
    \end{tabular}%
  }
  \label{tab:cls_datasets}%
\end{table} 

\begin{table}[!t]
  \centering
  \caption{Scene classification results of the five methods for seven datasets. OA is used as the evaluation index.}
    \resizebox{\linewidth}{!}{
      \begin{tabular}{c|c|ccccccc|c}
      \toprule
            & Method & \multicolumn{1}{c}{AID} & \multicolumn{1}{c}{NR} & \multicolumn{1}{c}{RSD46} & \multicolumn{1}{c}{PatternNet} & \multicolumn{1}{c}{UCMerced} & \multicolumn{1}{c}{EuroSAT} & \multicolumn{1}{c|}{NaSC-TG2} & \multicolumn{1}{c}{Mean} \\
      \midrule
      \multicolumn{1}{c|}{\multirow{5}[2]{*}{5 samples}} & Random initialization & 36.50 & 25.44 & 11.08 & 33.40 & 17.71 & 43.81 & 45.82 & 30.54 \\
            & ImageNet pretraining & 71.60 & 58.13 & 33.34 & 72.06 & \textbf{66.57} & 65.61 & 84.05 & 64.48 \\
            & SimCLR & 69.95 & 55.59 & 27.91 & 62.24 & 52.19 & 70.31 & 82.20 & 60.06 \\
            & MoCov2 & 73.30 & 63.16 & 30.82 & 66.04 & 58.19 & 72.44 & 83.02 & 63.85 \\
            & Our method & \textbf{78.55} & \textbf{62.89} & \textbf{33.76} & \textbf{73.68} & 61.52 & \textbf{74.70} & \textbf{85.57} & \textbf{67.24} \\
      \midrule
      \multicolumn{1}{c|}{\multirow{5}[2]{*}{20 samples}} & Random initialization & 57.85 & 47.27 & 20.35 & 45.58 & 33.14 & 65.39 & 64.65 & 47.75 \\
            & ImageNet pretraining & 81.65 & 71.92 & \textbf{47.36} & \textbf{77.50} & \textbf{75.24} & 79.70 & \textbf{92.67} & 75.15 \\
            & SimCLR & 78.65 & 70.41 & 37.83 & 67.02 & 64.76 & 83.15 & 89.10 & 70.13 \\
            & MoCov2 & 83.05 & 76.92 & 42.98 & 72.85 & 68.00 & 84.02 & 91.22 & 74.15 \\
            & Our method & \textbf{84.95} & \textbf{76.60} & 45.55 & 76.02 & 70.48 & \textbf{87.35} & 92.27 & \textbf{76.17} \\
      \midrule
      \multicolumn{1}{c|}{\multirow{5}[2]{*}{50 samples}} & Random initialization & 65.05 & 68.48 & 26.82 & 59.90 & 54.57 & 73.87 & 84.12 & 61.83 \\
            & ImageNet pretraining & 86.10 & 78.98 & 52.72 & 84.16 & \textbf{85.90} & 84.61 & \textbf{95.47} & 81.13 \\
            & SimCLR & 81.70 & 78.38 & 44.12 & 76.71 & 77.90 & 85.93 & 92.15 & 76.70 \\
            & MoCov2 & 85.40 & 83.33 & 49.13 & 79.80 & 79.81 & 86.98 & 94.05 & 79.79 \\
            & Our method & \textbf{88.55} & \textbf{83.57} & \textbf{52.96} & \textbf{85.38} & 83.81 & \textbf{89.54} & 95.05 & \textbf{82.69} \\
      \midrule
      \multicolumn{1}{c|}{\multirow{5}[2]{*}{100 samples}} & Random initialization & 77.15 & 61.17 & 35.48 & 71.27 & 61.24 & 79.61 & 92.02 & 68.28 \\
            & ImageNet pretraining & 85.65 & 82.51 & 56.56 & 87.38 & \textbf{86.00} & 86.69 & \textbf{96.47} & 83.04 \\
            & SimCLR & 84.80 & 82.62 & 49.89 & 83.63 & 77.90 & 88.24 & 94.50 & 80.23 \\
            & MoCov2 & 87.10 & 87.22 & 53.87 & 87.04 & 79.81 & 89.31 & 95.97 & 82.90 \\
            & Our method & \textbf{88.80} & \textbf{87.21} & \textbf{58.00} & \textbf{89.57} & 83.71 & \textbf{91.07} & 96.37 & \textbf{84.96} \\
      \bottomrule
      \end{tabular}%
    }
  \label{tab:cls_results}%
\end{table} 

\textbf{Task and dataset description.} Seven datasets shown in Table \ref{tab:cls_datasets} were used to evaluate the generalization capabilities of TOV model on scene classification task: Aerial Image Dataset (AID) \cite{Xia_Hu_Hu_Shi_Bai_Zhong_Zhang_Lu_2017}, NWPU-RESISC45 (NR) \cite{Cheng_Han_Lu_2017}, RSD46 \cite{Xiao_Long_Li_Wei_Tang_Liu_2017}, PatternNet \cite{Zhou_Newsam_Li_Shao_2018}, UC Merced \cite{Yang_Newsam_2010}, EuroSAT \cite{Helber_Bischke_Dengel_Borth_2019} and NaSC-TG2 \cite{Zhou_Li_Wu_Guo_Li_Xia_Zhao_2021}. For all the dataset, only RGB channels were used in the experiment. The overall accuracy (OA) was used to assess the performance.

\textbf{Fine-tuning settings.} For the proposed method and two compared SSL method, we add a simple fully-connected layer at the end of TOV model as scene classification adapter, and then fine-tune the model using 5, 20, 50, 100 labeled samples per category, respectively. During training, we used the Adam optimizer with a batch size of 32. The learning rate was initially set to be 0.001 and was reduced in a cosine manner within 200 epochs.

\textbf{Results and analysis.} Table \ref{tab:cls_results} shows the experiment results. The best results are marked in bold. From the results, we can get the following two findings:

First, our method consistently outperforms all compared methods in average of seven test datasets no matter how many samples are used for fine-tuning. For example, when using 5 samples per category for fine-tuning, the proposed method achieved a 4.3\% performance improvement in average compared to the second-best method. This result indicates the impressive generalization capabilities of TOV model on scene classification task.

Second, though ImageNet pretraining method outperforms the two SSL method in average, the advantage is not obvious. For example, when using 100 samples per category for fine-tuning, MoCov2 can achieve comparable results to the ImageNet pretraining method. Considering that obtaining the ImageNet pretraining model requires tens of millions labeled data for supervised learning, training a general model in a label-free and task-independent way is more effective and robust for RSIU tasks.

\subsubsection{Object detection}\label{subsubsec:object_detection}
\textbf{Task and dataset description.} Two datasets DOTA \cite{Xia_Bai_Ding_Zhu_Belongie_Luo_Datcu_Pelillo_Zhang_2018} and Levir \cite{Zou_Shi_2018} were used to evaluate the generalization capabilities of TOV model on object detection task, and the mean average precision (mAP50) is used to assess the performance.

\begin{table}[!t]
  \centering
  \caption{Object detection results of two datasets. mAP50 is used as the evaluation index.}
  \resizebox{0.7\textwidth}{!}{
    \begin{tabular}{c|c|cc|c}
    \toprule
    \multicolumn{1}{c|}{Proportion} & \multirow{2}[2]{*}{Method} & \multicolumn{1}{c}{\multirow{2}[2]{*}{DOTA}} & \multicolumn{1}{c|}{\multirow{2}[2]{*}{Levir}} & \multicolumn{1}{c}{\multirow{2}[2]{*}{Mean}} \\
    \multicolumn{1}{c|}{of training data} & \multicolumn{1}{c|}{} &       &       &  \\
    \midrule
    \multirow{5}[2]{*}{0.5\%} & Random initialization & 0.9   & 3.6   & 2.3 \\
          & ImageNet pretraining & 4.8   & \textbf{12.5} & 8.7 \\
          & SimCLR & 5.0   & 10.6  & 7.8 \\
          & MoCov2 & 3.2   & 10.5  & 6.9 \\
          & TOV   & \textbf{7.2} & 10.9  & \textbf{9.1} \\
    \midrule
    \multirow{5}[2]{*}{1.0\%} & Random initialization & 4.5   & 3.4   & 4.0 \\
          & ImageNet pretraining & 7.9   & 11.0  & 9.5 \\
          & SimCLR & \textbf{11.9} & 11.7  & 11.8 \\
          & MoCov2 & 10.6  & 10.1  & 10.4 \\
          & TOV   & 11.7  & \textbf{12.8} & \textbf{12.3} \\
    \midrule
    \multirow{5}[2]{*}{5.0\%} & Random initialization & 17.4  & 8.9   & 13.2 \\
          & ImageNet pretraining & 22.5  & 27.8  & 25.2 \\
          & SimCLR & 25.6  & 22.2  & 23.9 \\
          & MoCov2 & 26.1  & 25.5  & 25.8 \\
          & TOV   & \textbf{26.1} & \textbf{30.4} & \textbf{28.3} \\
    \bottomrule
    \end{tabular}%
  }
  \label{tab:det_results}%
\end{table} 

\textbf{1) DOTA} consists of RGB images and grayscale images. The RGB images are from Google Earth and CycloMedia while the grayscale images are from the panchromatic band of GF-2 and JL-1 satellite images. This dataset contains 188,282 objects from 15 categories.

\textbf{2) Levir} is collected from Google Earth and consists of over 22,000 images with a size of $800\times 600$ and the resolution of 0.2 $\mathrm{\sim}$ 1.0 m/pixels. It has three categories: airplane, ship and oil-tank.

\textbf{Fine-tuning settings}. For the proposed method and two compared SSL method, we joint use region proposal network and ROIHead in Faster RCNN\footnote{Faster RCNN was implemented by using MMDetection (https://github.com/open-mmlab/mmdetection).} as object detection adapter and then fine-tune the model using 0.5\%, 1.0\%, 5.0\% labeled samples of the whole dataset. During the training, we use the SGD optimizer with a batch size of 4 for fine-tuning. The learning rate was initially set as 0.0025 and was reduced in a cosine manner within 200 epochs.

\textbf{Results and analysis.} Table \ref{tab:det_results} shows the experiment results. The best results are marked in bold. Similar result can also be seen that our method consistently outperforms all compared methods in average of two test datasets no matter how many samples are used for fine-tuning. Moreover, we can observe that the performance of MoCov2 is better than ImageNet pretraining method when use 1.0\% and 5.0\% labeled data for fine-tuning, which further suggests the advantage of training TOV model in an SSL mechanism for RSIU.

\subsubsection{Semantic segmentation}\label{subsubsec:segmentation}
\textbf{Task and dataset description.} Three datasets DLRSD \cite{Shao_Yang_Zhou_2018}, DGLCC \cite{Demir_Koperski_Lindenbaum_Pang_Huang_Basu_Hughes_Tuia_Raskar_2018} and Potsdam dataset \cite{Rottensteiner_Sohn_Jung_Gerke_Baillard_Benitez_Breitkopf_2012} were used to evaluate the generalization capabilities of the proposed TOV model on semantic segmentation task, and the Mean Intersection over Union (MIoU) was used to assess the performance.

\textbf{1) DGLCC} is collected from DeepGlobe satellite, and contains 803 images with a size of $2448 \times 2448$ and the resolution of 0.5 m. The dataset is annotated in 7 classes.

\textbf{2) DLRSD} is a densely labeled dataset that consists of 2,100 RGB images with the size of $256 \times 256$ and the resolution of 0.3 m. The dataset is annotated in 17 classes.

\textbf{3) Potsdam} contains 38 UAV images with a size of $6000 \times 6000$ and the spatial resolution of 0.05 m. The dataset is annotated in 7 classes.

\textbf{Fine-tuning settings.} For the proposed method and two compared SSL method, we add the decoder in FCN \cite{longFullyConvolutionalNetworks2015} at the end of the encoder part of TOV model as an adapter for semantic segmentation tasks, and then fine-tune the model using 0.5\%, 1.0\%, 5.0\% labeled samples of the whole dataset. During the training, we use the Adam optimizer with a batch size of 32. The learning rate was initially set as 0.001 and was reduced in a cosine manner within 200 epochs.

\begin{table}[!t]
  \centering
  \caption{Semantic segmentation results of three datasets. MIoU is used as the evaluation index.}
  \resizebox{0.8\textwidth}{!}{
    \begin{tabular}{c|c|ccc|c}
    \toprule
    \multicolumn{1}{c|}{Proportion} & \multirow{2}[2]{*}{Method} & \multicolumn{1}{c}{\multirow{2}[2]{*}{DGLCC}} & \multicolumn{1}{c}{\multirow{2}[2]{*}{DLRSD}} & \multicolumn{1}{c|}{\multirow{2}[2]{*}{Potsdam}} & \multicolumn{1}{c}{\multirow{2}[2]{*}{Mean}} \\
    \multicolumn{1}{c|}{of training data} & \multicolumn{1}{c|}{} &       &       &       &  \\
    \midrule
    \multirow{5}[2]{*}{0.5\%} & Random initialization & 26.53 & 4.29  & 38.86 & 23.23 \\
          & ImageNet pretraining & 29.30 & 3.41  & 38.59 & 23.77 \\
          & SimCLR & \textbf{36.00} & 4.51  & 42.18 & 27.56 \\
          & MoCov2 & 27.50 & 4.44  & 39.69 & 23.88 \\
          & Our method & 35.86 & \textbf{6.90} & \textbf{50.23} & \textbf{31.00} \\
    \midrule
    \multirow{5}[2]{*}{1.0\%} & Random initialization & 27.55 & 7.93  & 38.15 & 24.54 \\
          & ImageNet pretraining & 33.47 & 12.44 & 39.77 & 28.56 \\
          & SimCLR & 34.71 & 13.17 & 40.53 & 29.47 \\
          & MoCov2 & 31.42 & 13.97 & 40.40 & 28.60 \\
          & Our method & \textbf{41.81} & \textbf{17.83} & \textbf{49.15} & \textbf{36.26} \\
    \midrule
    \multirow{5}[2]{*}{5.0\%} & Random initialization & 39.01 & 22.60 & 51.56 & 37.72 \\
          & ImageNet pretraining & 39.65 & 29.76 & 51.79 & 40.40 \\
          & SimCLR & 39.74 & 30.41 & 51.64 & 40.60 \\
          & MoCov2 & 38.44 & 31.12 & 51.75 & 40.44 \\
          & Our method & \textbf{41.31} & \textbf{39.29} & \textbf{60.34} & \textbf{46.98} \\
    \bottomrule
    \end{tabular}%
  }
  \label{tab:seg_results}%
\end{table} 

\textbf{Results and analysis.} Table \ref{tab:seg_results} shows the experiment results. The best results are marked in bold. Experimental results demonstrate that our method outperforms other comparison methods in most cases. Moreover, we can observe that ImageNet pretraining method does not significantly improve the semantic segmentation accuracy compared with random initialization, while the SSL learning method like SimCLR can outperform ImageNet pretraining in most cases. The main reason could be that ImageNet pretrained model is learned in a task-dependent way (i.e., scene classification task). Since the task of scene classification is not directly related to the task of semantic segmentation, resulting the generalization ability of learned features is not as good as that learned in a task-independent way.

\section{Discussion}\label{sec:dis}
In this section, we conduct a series of comparative experiments to further analyze the effect of two key factors on the performance of building the TOV model, including the influence of using different data sampling methods and the selection of learning paths during self-supervised optimization.

\subsection{Data sampling methods}\label{subsec:dis_sampling}
Rich and high-quality data is important to training TOV model for RSIU. Though this model can be trained in a label-free SSL way, it can hardly learn valuable remote sensing visual knowledge from an unlabeled dataset that contains large amounts of semantically meaningless content or has severely class-imbalanced distribution. To figure out how different sampling methods affect the performance of the constructed TOV models, we repeat the process of building TOV model by employing one of the following sampling methods each time.

\begin{table}[!t]
  \centering
  \caption{Analysis of different sampling methods for the performance of TOV model on the scene classification task. Five samples per class is used for fine-tuning. OA is used as the evaluation index. Best results are marked by bold font.}
  \resizebox{\textwidth}{!}{
    \begin{tabular}{c|ccccccc|c}
    \toprule
    Pretrained dataset & \multicolumn{1}{c}{AID} & \multicolumn{1}{c}{NR} & \multicolumn{1}{c}{RSD46} & \multicolumn{1}{c}{PatternNet} & \multicolumn{1}{c}{UCMerced} & \multicolumn{1}{c}{EuroSAT} & \multicolumn{1}{c|}{NaSC-TG2} & \multicolumn{1}{c}{Mean} \\
    \midrule
    TOV-RS-gridsampling & 63.76 & 48.06 & 24.74 & 64.74 & 45.31 & 73.24 & 82.04 & 57.42 \\
    TOV-RS-imbalanced & 71.05 & 49.66 & 28.55 & 63.25 & 51.46 & 69.32 & 82.07 & 59.33 \\
    TOV-RS-balanced & \textbf{78.55} & \textbf{62.89} & \textbf{33.76} & \textbf{73.68} & \textbf{61.52} & \textbf{74.70} & \textbf{85.57} & \textbf{67.24} \\
    \bottomrule
    \end{tabular}%
  }
  \label{tab:dis_sampling}%
\end{table} 

1) Grid sampling method. Give the set $H=\{\boldsymbol{I}_{1}, \boldsymbol{I}_{2}, \ldots, \boldsymbol{I}_{5000}\}$ of RSIs, we meshed each RSI $I_i$ in $H$ into $n_i$ non-overlapping patches with a size of $600\times 600$ pixels, and then randomly sample 600 patches. Finally, a dataset TOV-RS-gridsampling containing 3 million samples was obtained.

2) The proposed sampling method without resampling strategy. We use the data sampling method descried in Sec. \ref{subsubsec:sampling_natural} and \ref{subsubsec:sampling_man_made} for obtaining a class-imbalanced dataset, TOV-RS-imbalanced, containing 3 million samples.

3) The proposed sampling method with resampling strategy. We use the data sampling method descried in Sec \ref{subsec:data_acquisition} for obtaining a relative class-balanced dataset, TOV-RS-balanced, containing 0.5 million samples.

We evaluated the generalization capabilities of TOV model trained by different datasets on scene classification task. As shown in Table \ref{tab:dis_sampling}, though the size of dataset TOV-RS-gridsampling and TOV-RS-imbalanced is much larger than TOV-RS-balanced, TOV model learned from TOV-RS-balanced dataset significantly outperforms those learned from other two datasets for all seven datasets, with an average OA improvement of 17.1\% and 13.3\%. This result suggests that it is crucial to choose an appropriate sampling method to obtain high-quality dataset for training TOV model for RSIU. This result may be from two reasons:

First, the grid sampling approach may sample large amounts of semantically meaningless data, which confuses the feature representation learning of TOV model.

Second, the problem of data imbalance existing in the first and second data sampling methods poses challenges in training TOV using contrastive SSL method. The idea of contrastive SSL learning is to encourage a model to learn invariance features by distinguishing between positive and negative samples. Since there is no annotation information in self-supervised learning, false negative samples (i.e., samples belonging to the same class) are more likely to be sampled in class-imbalanced datasets. As a result, the more imbalanced the classes are, the more false negative samples are sampled. This phenomenon potentially let TOV model push features that belong to the same class farther away, and thus hurts the model’s performance.

\begin{table}[!t]
  \centering
  \caption{Analysis of the selection of learning paths for the performance of TOV model on scene classification task. Five samples per class is used for fine-tuning. OA is used as the evaluation index. Best results are marked by bold font.}
  \resizebox{\textwidth}{!}{
  \begin{tabular}{cc|ccccccc|c}
    \toprule
    \multirow{2}[2]{*}{Learning path} & \multicolumn{1}{c|}{Memory} & \multirow{2}[2]{*}{AID} & \multirow{2}[2]{*}{NR} & \multirow{2}[2]{*}{RSD46} & \multirow{2}[2]{*}{PatternNet} & \multirow{2}[2]{*}{UCMerced} & \multirow{2}[2]{*}{EuroSAT} & \multicolumn{1}{c|}{\multirow{2}[2]{*}{NaSC-TG2}} & \multirow{2}[2]{*}{Mean} \\
     & \multicolumn{1}{c|}{maintenance} &  &  &  & &  &  & \multicolumn{1}{c|}{} &  \\
    \midrule
    $\langle D_{NI}\rangle $     & -     & 69.65 & 57.35 & 31.73 & 69.21 & 56.95 & 73.26 & 83.30 & 63.06 \\
    $\langle D_{RS}\rangle $     & -     & 69.95 & 55.59 & 27.91 & 62.24 & 52.19 & 70.31 & 82.20 & 60.06 \\
    $\langle D_{NI}\mathrm{,\ }D_{RS}\rangle $ & w/o   & 68.60 & 56.05 & 27.34 & 63.47 & 50.95 & 72.74 & 80.12 & 59.90 \\
    $\langle D_{NI}\mathrm{,\ }D_{RS}\rangle $ & w/    & \textbf{78.55} & \textbf{62.89} & \textbf{33.76} & \textbf{73.68} & \textbf{61.52} & \textbf{74.70} & \textbf{85.57} & \textbf{67.24} \\
    \bottomrule
    \end{tabular}%
  }
  \label{tab:dis_learning_path}%
\end{table} 

\subsection{The selection of learning paths}\label{subsec:dis_learning_path}
During self-supervised optimization, we design a human-like learning path, which first learns general knowledge from web-scale natural images and then learns domain-relevant specialized knowledge from unlabeled RSIs. To figure out how learning path selection affects the performance of TOV model for RSIU, we designed comparison experiments as shown in Table \ref{tab:dis_learning_path}, where $\langle D_{NI}\rangle $ and $\langle D_{RS}\rangle $ means only using natural image dataset TOV-NI and the RSI dataset TOV-RS-balanced respectively for learning TOV model, $\langle D_{NI}\mathrm{,\ }D_{RS}\rangle $ represents using both TOV-NI and TOV-RS-balanced along a general to specialized learning path for learning TOV model. From the results shown in Table \ref{tab:dis_learning_path}, we found two phenomena.

Firstly, training TOV model along a learning path from general knowledge to specialized knowledge can improve model performance greatly, with an average OA improvement of 6.63\% and 11.95\% compared with that only using natural image dataset or RSI dataset, respectively. The main reason could be that different types of datasets can provide complementary knowledge. Besides, the model trained only using RSI dataset is even worse than the model trained only using nature image dataset, which confirm that remote sensing data are not sufficient to support TOV model to learn generic visual representations.

Second, designing a memory maintenance strategy for continual learning on different kinds of dataset is important, because TOV model has the problem of catastrophic forgetting which occurs frequently when the network is trained sequentially on different datasets. If training TOV model without using the memory maintenance strategy, it may totally forget what it has learned during training and get similar results as only using one dataset. (See row 2 and row 3 of Table \ref{tab:dis_learning_path}).

\section{Conclusions}\label{sec:conclusion}
In this study, we give the definition of TOV model for RSIU, and investigate a new paradigm for training TOV. Moreover, we perform comprehensive comparative study by analyzing two key factors on the performance of building TOV model for RSIU, including the influence of using different data sampling methods and the selection of learning paths during self-supervised optimization. By combining our findings, our TOV model has shown impressive generalization capabilities across various RSIU tasks and outperforms dominant ImageNet supervised pretrained method as well as two recently proposed SSL pretrained methods on majority of 12 publicly available benchmarks. Our future work aims at building TOV 2.0 model for RSIU considering both task granularity and data modality. We expect TOV 2.0 to be broadly adaptable to multiple RSIU tasks and data modalities like hyperspectral image, SAR, and even video data, which can potentially pave the way for building general intelligence in the remote sensing field.

\section*{Acknowledgment}\label{sec:acknowlegements}
The work presented in this paper was supported by the National Key Research and Development Program (grant number 2018YFB0504501); The National Natural Science Foundation of China (No. 42171376, 41771458, 41871364); The Natural Science Foundation of Hunan (2021JJ30815); The Young Elite Scientists Sponsorship Program by Hunan province of China (No. 2018RS3012); Hunan Science and Technology Department Innovation Platform Open Fund Project (18K005); and the High Performance Computing Center of Central South University.

\section*{References}

\bibliography{mybibfile}

\end{document}